\documentclass{article}

\usepackage{PRIMEarxiv}

\usepackage[utf8]{inputenc} 
\usepackage[T1]{fontenc}    
\usepackage{hyperref}       
\usepackage[numbers]{natbib}
\usepackage{url}            
\usepackage{booktabs}       
\usepackage{amsfonts}       
\usepackage{nicefrac}       
\usepackage{microtype}      
\usepackage{lipsum}
\usepackage{fancyhdr}       
\usepackage{graphicx}       
\graphicspath{{media/}}     
\usepackage{booktabs}
\usepackage{multirow}
\usepackage{makecell}
\usepackage{array}
\usepackage{amsmath}
\usepackage[linesnumbered, ruled, vlined]{algorithm2e}
\usepackage{longtable}
\usepackage{caption}
\usepackage{float} 
\usepackage{array,tabularx}
\usepackage{enumitem}

\title{Evolutionary Token-Level Prompt Optimization for Diffusion Models 
}

\author{
  Domício Pereira Neto\\
  University of Coimbra \\
  Coimbra, Portugal\\
  \texttt{dneto@dei.uc.pt} \\
   \And
  João Correia\\
  University of Coimbra \\
  Coimbra, Portugal\\
  \texttt{jncor@dei.uc.pt} \\
   \And
  Penousal Machado\\
  University of Coimbra \\
  Coimbra, Portugal\\
  \texttt{machado@dei.uc.pt} \\
}

\begin{document}
\maketitle

\begin{abstract}
Text-to-image diffusion models exhibit strong generative performance but remain highly sensitive to prompt formulation, often requiring extensive manual trial and error to obtain satisfactory results. This motivates the development of automated, model-agnostic prompt optimization methods that can systematically explore the conditioning space beyond conventional text rewriting. This work investigates the use of a Genetic Algorithm (GA) for prompt optimization by directly evolving the token vectors employed by CLIP-based diffusion models. The GA optimizes a fitness function that combines aesthetic quality, measured by the LAION Aesthetic Predictor V2, with prompt–image alignment, assessed via CLIPScore. Experiments on 36 prompts from the Parti Prompts (P2) dataset show that the proposed approach outperforms the baseline methods, including Promptist and random search, achieving up to a 23.93\% improvement in fitness. Overall, the method is adaptable to image generation models with tokenized text encoders and provides a modular framework for future extensions, the limitations and prospects of which are discussed.
\end{abstract}


\section{Introduction}
Prompt optimization involves the automatic refinement of textual input, either by rewriting the prompt itself or by optimizing a continuous representation derived from it, such as tokens or embedding vectors, which can be applied to different types of uses, such as text-to-image generation. The objective is to enhance the output according to the given preferences. As modern diffusion models exhibit high sensitivity to prompt phrasing and emphasis, minor modifications can result in significant variations in composition, style and semantic alignment \cite{Rombach2022}. Consequently, prompt optimization has become increasingly crucial for enhancing the reliability and controllability of image generation, thereby minimizing the manual trial-and-error process required to achieve aesthetically pleasing outputs that align with the user's intent.

Current research on prompt optimization encompasses a range of methodologies that balance interpretability, flexibility, and computational costs. A significant portion of these approaches operates within discrete search spaces, employing techniques such as prompt rewriting, expansion or editing \cite{hao2023, tran2023}. These techniques are driven by large language models (LLMs), learned "prompt improvers,” or rule-based heuristics to enhance  stylistic specificity while maintaining the original semantics. Concurrently, continuous optimization approaches conceptualize prompt conditioning as a differentiable control signal, exploring embeddings or adapter-like vectors to optimize objective functions \cite{clare2023, Wang2024, salvenmoser2025}, such as aesthetic predictors, CLIP-based alignment scoring, and rewards for downstream tasks. A central challenge across both approaches is achieving a balance between visual appeal and fidelity to the intended content within realistic constraints, as higher-quality optimization often necessitates numerous model evaluations and may overfit proxies that do not fully encapsulate human preferences.

In this context, considering the unexplored potential of discrete prompt search and the capabilities of Evolutionary Algorithms (EAs) in such complex spaces, we chose to study the application of a Genetic Algorithm (GA) to advance prompt optimization. By defining the search space as the token vocabulary from CLIP, we evolved a vector of tokens for Stable Diffusion XL Turbo (SDXL Turbo). This process aims to optimize a weighted fitness function that integrates the aesthetic score from the LAION Aesthetic Predictor V2 and prompt alignment from CLIPScore.

To assess the competitiveness of this approach, we selected Promptist as one of the baseline models, which is considered state-of-the-art in prompt optimization \cite{hao2023}. Because of its architecture, which is centered on an LLM, it can perform zero-shot prompt optimization. However, it has certain limitations, such as being specifically trained for a particular diffusion model (Stable Diffusion 1.4) and exhibiting training data bias from the Lexica dataset. In addition to Promptist, a random search was also used as a baseline. The experimental setup involved optimizing 36 prompts, corresponding to three of each of the 12 categories within the Parti Prompts (P2) dataset, with the objective of maximizing the fitness function, which is composed of both the aesthetic score and prompt alignment. 

Our findings indicate the superiority of the evolutionary approach over the baseline methods used in the experiments. The GA is effective in enhancing both aesthetics and prompt-image alignment, particularly when the population is initialized with mutations of the original prompt. Beyond performance, a notable advantage of our approach is its adaptability to any method that utilizes the CLIP text encoder or any other text encoder that incorporates tokenization into its pipeline.

The contributions of this work are as follows: (i) A comprehensive examination of GA within the field of prompt optimization, specifically applied to enhance the prompt token vector of a text-to-image diffusion model, (ii) the GA prompt optimization algorithm\footnote{\url{https://github.com/domiciopereiraneto/evo_tokens}} has been made publicly accessible to facilitate the replication of results and to encourage further experimentation by the community, and (iii) a discussion of the limitations encountered and the identification of future research avenues, including proposed strategies for developing a fully modular token optimization framework.

The remainder of this paper is organized as follows. In Section \ref{sec:related} the current literature is discussed. Section \ref{sec:approach} provides a detailed explanation of the methodology employed in this work. Section \ref{sec:setup} details the experimental setup, followed by the presentation of the experimental results in Section \ref{sec:results}. Finally, the conclusions are presented in Section \ref{sec:conclusions}.

\section{Related Work}
\label{sec:related}

Prompt optimization for text-to-image generation has garnered significant attention because of the sensitivity of diffusion models to prompt phrasing \cite{dehouche2023, khan2025}. Current methodologies can be categorized into discrete text space techniques, such as rewriting using LLMs or employing heuristics, and continuous optimization approaches, which involve adjusting embeddings or adapter vectors to optimize objectives such as aesthetics or semantic alignment. Challenges persist in achieving a balance between visual quality and content fidelity, while managing computational costs.

In the domain of prompt optimization utilizing LLMs, Promptist has established a benchmark through its framework, which modifies user prompts using the GPT-2 model \cite{hao2023}. This model was fine-tuned with reinforcement learning, employing the Lexica dataset and Stable Diffusion 1.4 to enhance CLIP-based aesthetics and prompt alignment \cite{shen2024}. Consequently, the primary advantage of Promptist is its capability as a zero-shot, black-box, and lightweight prompt optimizer. However, it is also subject to biases inherent in the training data, costly to retrain, and is specific to one diffusion model.
 
In another perspective, Promptify employs an interactive methodology by utilizing LLMs to assist users in refining and exploring text prompts for text-to-image generation \cite{brade2023}. The system incorporates a feedback loop that enables users to organize generated images and receive targeted recommendations for prompt enhancement based on their preferences. Although user-friendly and model-agnostic, its efficacy is contingent on the quality of the LLMs employed.

Reliance on LLMs may constrain the solution space to the knowledge embedded within these models. Consequently, heuristics and metaheuristics may offer the potential to discover solutions that are inaccessible through conventional human vocabularies, syntax, and semantics. In this context, Evolutionary Algorithms (EAs) utilize iterative processes inspired by biological evolution, such as selection, mutation, and recombination, to effectively search complex and high-dimensional solution spaces, enabling the identification of innovative and optimized solutions beyond the limitations of fixed linguistic frameworks \cite{bartz2014}.

In the work by \citeauthor{tran2023}, the researchers introduced a novel GA implementation named Population–Offspring–Population (POPOP) \cite{tran2023}. This development primarily aimed to address the shortcomings identified in EvoGen (another GA prompt optimizer) \cite{petersen2022}, such as the absence of elitism in selection and inefficiencies in aligning with the target styles. The fitness function is constructed as a weighted sum of the CLIP Aesthetic Model score and the cosine distance loss. In this work, the weights were set at 0.4 for aesthetics and 0.6 for style alignment.

Beyond text evolution, there is also embedding optimization. In the latter, the latent space of text representation is explored and manipulated to achieve the desired outcomes. An example of this is presented in the work of \citeauthor{salvenmoser2025}, who introduced a GA-based framework that directly evolves the prompt embedding vectors used to condition text-to-image diffusion models, thereby circumventing traditional discrete prompt engineering \cite{salvenmoser2025}. This type of approach offers a gradient-free method to explore the complex and high-dimensional search space of prompt text embeddings. 

Based on the prior research, we conclude that directly evolving text strings may be less effective in exploring the search space compared to optimizing prompt embeddings. However, the latter involves higher-dimensional spaces, which can result in significant computational costs. As an intermediary approach between text strings and embeddings, we have developed a GA to evolve prompt tokens. These tokens are the fundamental textual units generated through tokenization, enabling the encoder to transform text into numerical representations that align with images within a shared embedding space. This methodology was tested using SDXL Turbo, with image evaluation conducted through a fitness function that integrates both an aesthetic score and an image-prompt alignment score. The objective is to achieve deeper optimization capabilities through token evolution, thereby improving both the aesthetic quality and semantic alignment.

\section{Methodology}
\label{sec:approach}

\begin{figure}[htbp]
  \centering
  \includegraphics[width=16cm]{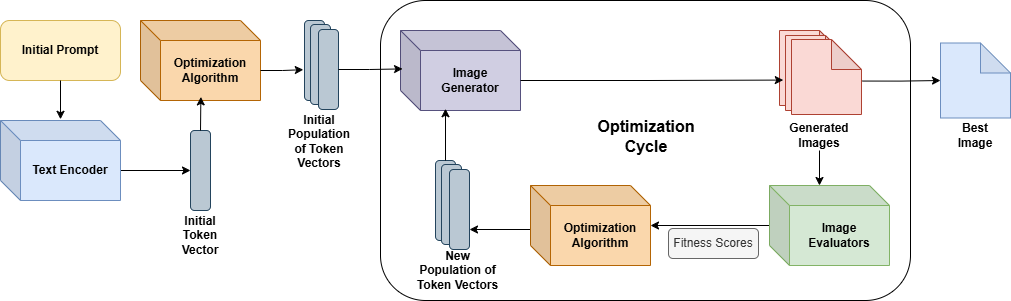}
  \caption{General structure and workflow of the evolutionary token optimization studied in this work. The main components and their respective inputs and outputs are depicted.}
  \label{fig:framework}
\end{figure}

This work evaluated the performance of a GA in refining prompt token vectors for image generation using diffusion models. In this context, tokens refer to discrete units of text, which are often entire words or subword segments generated by a tokenizer during prompt processing. CLIP, one of the most common text encoders, initially tokenizes the prompt into token IDs, associates each token with a learned embedding vector, and subsequently contextualizes these vectors through a transformer to yield a final latent text representation. These token vectors constitute the numerical form of the prompt that conditions the diffusion models. Thus, optimizing token vectors serves as a high-level approach to directly adjusting the embedding signal of a prompt, rather than simply changing its raw text. Figure \ref{fig:framework} presents the architecture of Stable Diffusion, including the prompt tokenization process. 

The process of evolutionary token optimization is as follows: the text encoder tokenizes the provided prompt. Three distinct methods are utilized for population initialization: (i) mutations of the initial token vector, (ii) empty vectors, and (iii) randomly generated token vectors. This initial population subsequently enters the optimization cycle, wherein the image generator (SDXL Turbo) is guided by each token vector to produce the images. These images are assessed based on aesthetics and prompt-image alignment, resulting in fitness scores that steer the GA in selecting the most suitable individuals for reproduction in subsequent generations.

In this section, the components shown in Figure \ref{fig:framework} are presented in detail.

\subsection{Image Generation \& Text Encoding}

Currently, there is a wide array of open-source image generation models, with sizes ranging from a few million parameters to several hundred billion. Among the most advanced models are diffusion models, with examples such as DeepFloyd-IF, which involves three diffusion stages and a large T5-family text encoder \cite{deepfloyd2023if}, DiT/PixArt-style diffusion transformers \cite{chen2023}, and SDXL pipelines that employ a base and refiner pair with numerous sampling steps to achieve a resolution of 1024$\times$1024 \cite{podell2023}. Although these models are precise, they can be expensive to run because of the need for numerous sequential denoising steps, multiple UNets or upsamplers, and large text encoders, all of which increase FLOPs, memory usage, and latency while reducing the batch size. Consequently, we opted for the well-known SDXL Turbo \cite{sauer2025}, a distilled version of the SDXL pipeline, which can generate high-quality images with only 1 to 4 denoising steps, as opposed to the approximately 50 steps required by the standard SDXL.

Text encoders are frequently built into image generation models, with CLIP (Contrastive Language-Image Pre-training) being particularly noteworthy because to its robust cross-modal capabilities, open accessibility, and ease of integration \cite{radford2021}, thereby establishing it as a standard for conditioning text-to-image generation models. Other notable text encoders include T5, which offers rich contextual embeddings for generative conditioning \cite{raffel2020}. The SDXL Turbo model employs CLIP.

\subsection{Image Evaluation}

We have selected an image evaluation methodology that integrates aesthetic evaluation with prompt-image alignment. The models utilized for this purpose are delineated below.

The LAION Aesthetic Predictor V2, created by the LAION community, is a straightforward regressor that evaluates the aesthetic quality of images as perceived by humans on a scale of 1 to 10 \cite{Schuhmann2022}. Its original purpose was to help select better subsets from large web datasets, such as LAION-5B, and to support training and evaluation processes that benefit from a quick, automated aesthetic rating. In our work, it functions as part of the fitness function to ensure a balance between visual appeal and text-to-image alignment while maintaining the efficiency required for large-scale evaluation cycles.

For semantic evaluation, we chose CLIPScore, which is directly derived from OpenAI’s CLIP \cite{hessel2021}. The CLIP encoder generates an image embedding $f_I(x)$ and a text embedding $f_T(p)$; their cosine similarity results in

\begin{equation}
    \mathrm{CLIPScore}(x,p)=\frac{\langle f_I(x),f_T(p)\rangle}{\|f_I(x)\|\,\|f_T(p)\|},
\end{equation}

which assesses the alignment between prompts and images. Implementations might use temperature scaling or normalization, but the primary indicator is this similarity, ranging from $[-1,1]$.

CLIPScore is utilized for zero-shot classification, cross-modal retrieval, caption reranking, and as an evaluative or guiding tool in generative processes. It functions efficiently, necessitating only a single forward pass through each encoder per sample, thereby rendering it suitable for extensive sweeps. It is sensitive to factors such as prompt wording, length, and dataset bias. In our work, we compute the CLIPScore for each generated image and its corresponding prompt, subsequently integrating it with the LAION Aesthetic V2 score to ascertain the fitness employed by both sep-CMA-ES and Adam.

\subsection{Optimization Algorithms}

Genetic Algorithms (GAs) enable efficient searches over large, high-dimensional, and non-differentiable spaces without gradient information. Through population-based exploration and stochastic operators, GAs can escape local optima and find diverse solutions that deterministic methods cannot. This makes them effective for optimizing prompt representations, where token configurations and outputs have complex relationships requiring black-box evaluation.

We employed a GA to optimize a weighted combination of aesthetic quality and prompt-image alignment by refining a population of prompt token vectors as described below.

\noindent Let:
\begin{itemize}[itemsep=5pt, topsep=0pt, parsep=0pt, partopsep=0pt]
    \item $K$: number of tokens per vector;
    \item $\mathbf{Z} = [\mathbf{z}_1,\dots,\mathbf{z}_K]\in\mathbb{R}^{K\times d}$: \emph{prompt token-vector} individual (genotype), where each $\mathbf{z}_k\in\mathbb{R}^d$ is a token embedding to be optimized;
    \item $p$: fixed text prompt (used only in the CLIP-text branch);
    \item $G(\mathbf{Z})$: generative model producing an image $\mathbf{x}$ when conditioned on $\mathbf{Z}$ (e.g., via cross-attention);
    \item $S_{\text{aest}}(\mathbf{x})\in[1,10]$, $S_{\text{clip}}(\mathbf{x},p)\in[-1,1]$;
    \item $\hat S_{\text{aest}}(\mathbf{x})=\mathrm{norm}_a\!\left(S_{\text{aest}}(\mathbf{x})\right)\in[0,1]$;
    \item $\hat S_{\text{clip}}(\mathbf{x},p)=\mathrm{norm}_c\!\left(S_{\text{clip}}(\mathbf{x},p)\right)\in[0,1]$;
    \item $a,b\ge0$, $a+b=1$: metric weights;
    \item $N$: population size; $T$: number of generations.
\end{itemize}

For an individual $\mathbf{Z}$, the fitness functions are defined as:
\begin{equation}
    F(\mathbf{Z}) \;=\; a\,\hat S_{\text{aest}}\!\big(G(\mathbf{Z})\big)\;+\; b\,\hat S_{\text{clip}}\!\big(G(\mathbf{Z}),p\big).
    \label{eq:fitness}
\end{equation}

Let the population at generation $t$ be:
\begin{equation}
    \mathcal{P}^{(t)}=\{\mathbf{Z}^{(t)}_1,\dots,\mathbf{Z}^{(t)}_N\}, \qquad t=0,\dots,T.
\end{equation}

The GA produces the next population via selection, crossover, and mutation:
\begin{equation}
    \mathcal{P}^{(t+1)} \;=\; \mathrm{GA}\!\Big(\mathcal{P}^{(t)};\ \mathrm{Sel},\mathrm{Xover},\mathrm{Mut},\mathrm{Elite}\Big),
    \label{eq:ga_update}
\end{equation}

and the GA seeks a high-fitness individual over the course of evolution. The best-so-far solution is
\begin{equation}
    \mathbf{z}^* \;=\; \arg\max_{\mathbf{z}} \; F(\mathbf{z}).
    \label{eq:objective_ga}
\end{equation}

The used GA employed tournament selection with size $k$ to form the mating pool, choosing winners probabilistically by fitness. One-point crossover performs reproduction by exchanging subsequences between parent token vectors. Uniform integer mutation randomly replaces tokens with valid embedding indices at a given probability. Finally, elitism preserves top performers by copying them to the next generation. 

Three methods were used for population initialization. The first method, GA Mutated, creates individuals as mutated copies of the initial prompt token vector, applying the same uniform integer mutation used in the evolution cycle. The second method, GA Empty, initializes the population with vectors composed entirely of padding tokens that are ignored by the encoder. This approach favors solutions with fewer tokens (excluding padding) and potentially simpler prompts. The final method, GA Random, as the name suggests, populates the token vectors with random tokens, thereby offering a broad spectrum of initial points within the search space.

\section{Experimental Setup}
\label{sec:setup}

For this work, we selected the Parti Prompts (P2) dataset to sample test prompts. P2 comprises more than 1600 prompts categorized into 12 distinct groups: Abstract, Vehicles, Illustrations, Arts, World Knowledge, People, Animals, Artifacts, Food \& Beverage, Produce \& Plants, Outdoor Scenes, and Indoor Scenes. Running the optimization experiments on the entire dataset would demand several thousand GPU hours; therefore, a smaller subset of 36 prompts was picked with random sampling, following a proportion of three prompts for each category. These selected prompts reflected the following distribution of challenge types: Basic challenges (8), Fine-grained Detail (7), Simple Detail (5), Complex (5), Style \& Format (4), Imagination (2), Writing \& Symbols (2), Quantity (2), and Linguistic Structures (1).

The experiments involved running optimization algorithms for each of the 36 prompts. The GA was set to run for 100 generations. The parameters are outlined in Table \ref{tab:parameters}.

\begin{table}[htb]
    \centering
    \caption{Parameters used in the experimentation. The first set of parameters are related to SDXL Turbo, while the second is related to the GA.}
    \begin{tabular}{c|c}
    \hline
        Parameter & Value\\ \hline
        Inference steps & 1\\
        Guidance Scale & 0\\ 
        Image Size & 512x512 \\ \hline
        $(a,b)$ & $(0.4,0.6)$ \\
        Generations & 100 \\
        Population size & 64\\ 
        Crossover prob. & 0.7\\
        Mutation prob. & 0.9\\
        Ind. mutation prob. & 0.1\\
        Tournment size & 3\\ 
        Elitism factor & 1\\ \hline
    \end{tabular}
    \label{tab:parameters}
\end{table}

The initial three parameters listed in the table pertain to SDXL Turbo. We opted for a single inference step to leverage the rapid image generation required for the experiment, which involves creating thousands of images. The guidance scale and image dimensions were maintained at their default settings of 0 and 512 × 512, respectively. Moreover, as the fitness function is defined by a balance between the two evaluation metrics (Eq. \ref{eq:fitness}), we have defined a weight combination of $(a,b)=(0.4,0.6)$ for the experiment, which prioritizes prompt-image alignment improvement while still upholding advancements in aesthetics. The remaining GA parameters were manually tuned and may be subject to a specific study in the future.

In terms of baseline approaches, Promptist, a pre-trained prompt optimizer, was executed once for each prompt, using the same fitness function to evaluate the results. In contrast, random search, an iterative algorithm, was performed for the same number of evaluations as the GA, specifically $100 \text{ generations } \times 64 \text{ individuals}=6400$ evaluations or iterations.

\section{Experimental Results and Discussion}
\label{sec:results}

\begin{figure*}[htbp]
  \centering
  \includegraphics[width=\linewidth]{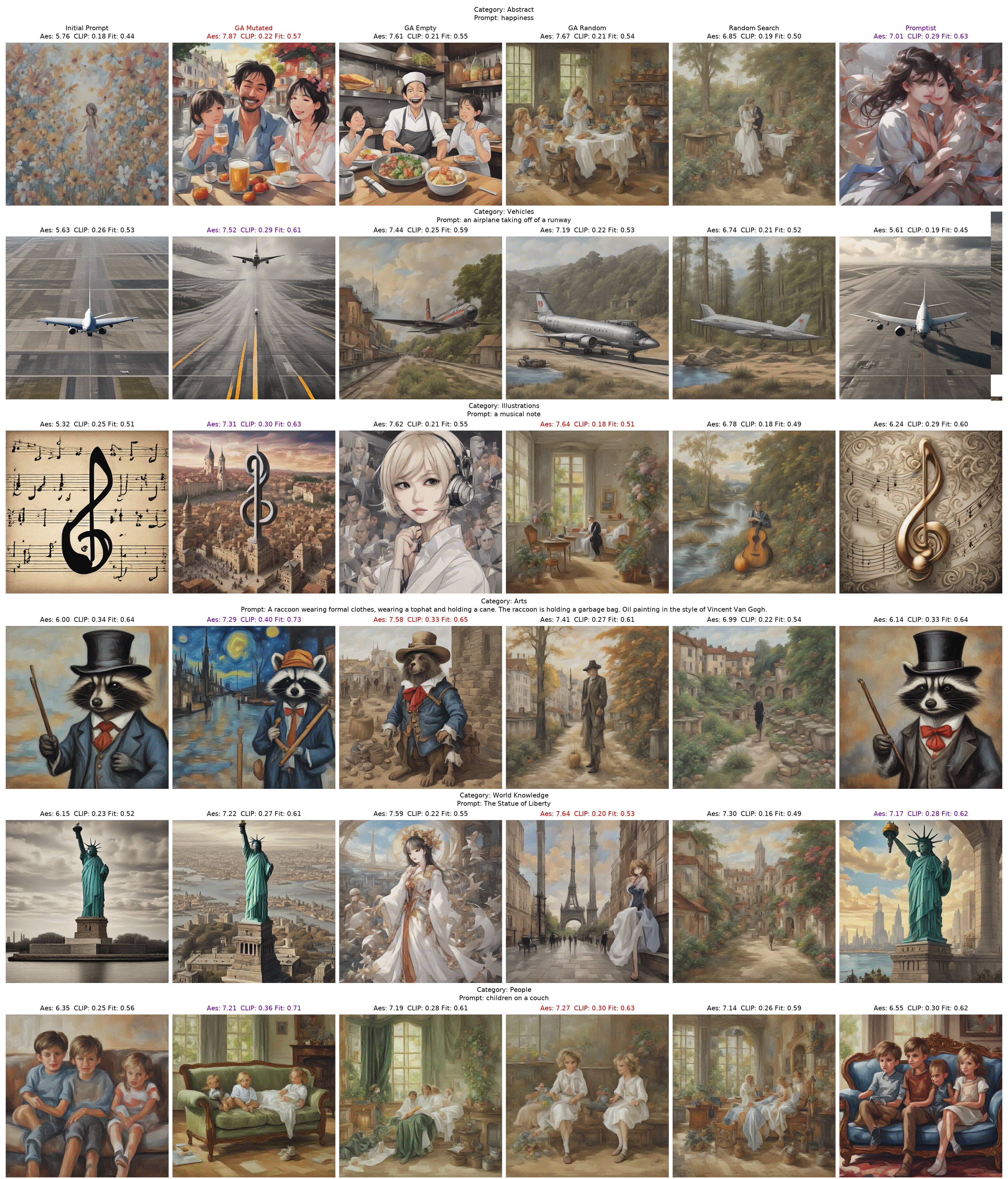}
  \caption{Final outputs from baseline SDXL Turbo, GA Mutated, GA Empty, GA Random, Random Search, and Promptist for 6 prompts of the first 6 P2 categories (see Section \ref{sec:setup}). Rows correspond to prompts and columns to methods, with aesthetic, CLIP, and fitness scores above each image; purple marks the highest-fitness image, while red or blue mark the best aesthetic or CLIPScore when they do not match the fitness optimum.}
  \label{fig:results_1}
\end{figure*}

\begin{figure*}[htbp]
  \centering
  \includegraphics[width=\linewidth]{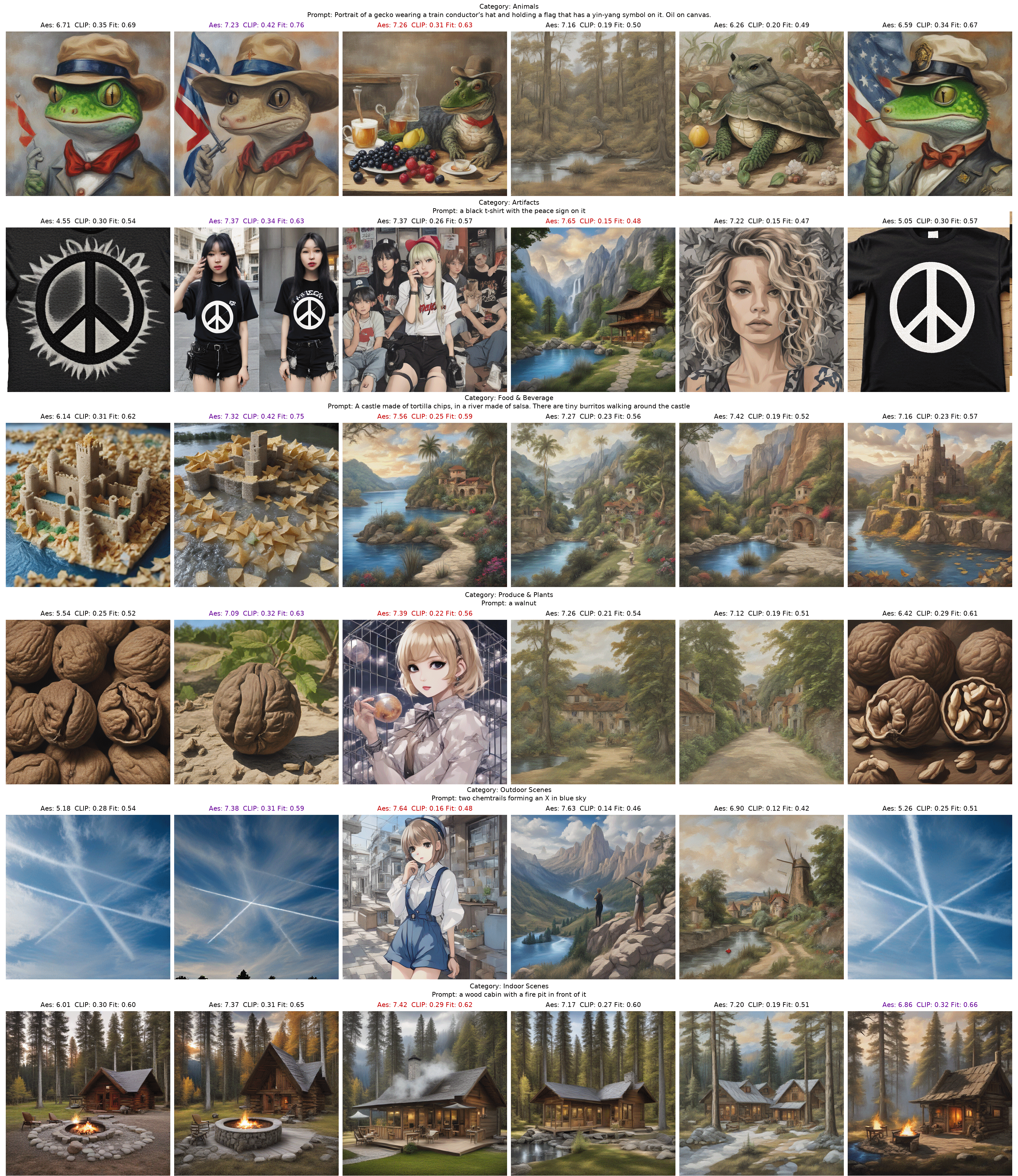}
  \caption{Final outputs from baseline SDXL Turbo, GA Mutated, GA Empty, GA Random, Random Search, and Promptist for 6 prompts of the last 6 P2 categories (see Section \ref{sec:setup}). Rows correspond to prompts and columns to methods, with aesthetic, CLIP, and fitness scores above each image; purple marks the highest-fitness image, while red or blue mark the best aesthetic or CLIPScore when they do not match the fitness optimum}
  \label{fig:results_2}
\end{figure*}

In this section, we present the results and discussion derived from the practical experiments. Initially, Table \ref{tab:results_ga} offers a detailed account of the quantitative results measured in terms of aesthetic score, CLIPScore, and fitness. Figures \ref{fig:results_1} and \ref{fig:results_2} illustrate a subset of the final outputs.

\begin{table*}[htbp]
\centering
\scriptsize
\setlength{\tabcolsep}{4.6pt}
\caption{Results comparison between SDXL Turbo with no optimization and prompt optimization methods under weights \((a,b)=(0.4,0.6)\). Each GA line corresponds to a different type of population initialization. The results are reported for LAION Aesthetic V2, CLIPScore, and Fitness. The columns show the mean, standard deviation, maximum, and percentage change relative to the original baseline. The highest mean and percentage change per metric are highlighted in bold, as well as the algorithm with highest average fitness.}
\begin{tabular}{lrrrrrrrrrrrrr}
\toprule
\multirow{2}{*}{Algorithm} &
\multicolumn{4}{c}{LAION Aesthetic V2 [1,10] $\uparrow$} &
\multicolumn{4}{c}{CLIPScore [-1,1] $\uparrow$} &
\multicolumn{4}{c}{Fitness [0,1] $\uparrow$} &
\multirow{2}{*}{Wins [0-36] $\uparrow$} \\
\cmidrule(lr){2-5}\cmidrule(lr){6-9}\cmidrule(lr){10-13}
& Avg. & Std. & Max & $\Delta$ base (\%) & Avg. & Std. & Max & $\Delta$ base (\%) & Avg. & Std. & Max & $\Delta$ base (\%) & \# prompts \\
\midrule
\makecell{SDXL Turbo\\(no optimization)}
& 5.78 & 0.56 & 6.81 & 0.00
& 0.2672 & 0.0482 & 0.3909 & 0.00
& 0.5519 & 0.0617 & 0.7311 & 0.00
& 0\\
\midrule
\textbf{GA Mutated}
& 7.30 & 0.20 & 7.87 & 26.29
& \textbf{0.3266} & 0.0522 & 0.4209 & \textbf{22.22}
& \textbf{0.6840} & 0.0596 & 0.7944 & \textbf{23.93}
& 28\\
GA Empty
& \textbf{7.45} & 0.21 & 7.95 & \textbf{28.94}
& 0.2562 & 0.0397 & 0.3752 & -4.12
& 0.6056 & 0.0444 & 0.7450 & 9.73
& 1\\
GA Random
& 7.39 & 0.23 & 7.91 & 27.88
& 0.2248 & 0.0381 & 0.3040 & -15.88
& 0.5654 & 0.0423 & 0.6572 & 2.45
& 0\\
\midrule
Random Search
& 6.93 & 0.37 & 7.43 & 19.87
& 0.1946 & 0.0343 & 0.2830 & -27.18
& 0.5107 & 0.0408 & 0.5947 & -7.47
& 0\\
Promptist
& 6.43 & 0.69 & 7.44 & 11.19
& 0.2808 & 0.0409 & 0.3428 & 5.09
& 0.5941 & 0.0552 & 0.6743 & 7.64
& 7\\
\bottomrule
\end{tabular}
\label{tab:results_ga}
\end{table*}

The results presented in Table \ref{tab:results_ga} show that among the GA variations, GA Empty demonstrated the highest performance in terms of aesthetic improvement, achieving an average score of 7.45 within the expected range of $(1,10)$. This was followed by GA Random with a score of 7.39 and GA Mutated with a score of 7.30. In comparison, among the baseline methods, Random Search attained a score of 6.93, whereas Promptist recorded the lowest score among the optimization methods, with 6.43. The standard SDXL Turbo yielded an average score of 5.78.

Regarding CLIPScore, which ranges from $(-1,1)$, the GA Mutated method achieved the highest average score of 0.3266. This was followed by Promptist (0.2808) and GA Empty (0.2562). Subsequently, GA Random obtained a score of 0.2248, whereas Random Search recorded the lowest score among the optimizers at 0.1946. SDXL Turbo without optimization had an average of 0.2672.

Consequently, the fitness scores, weighted at $0.4$ for aesthetics and $0.6$ for CLIPScore (refer to Eq. \ref{eq:fitness}), yielded the following results within the range of $(0,1)$: GA Mutated achieved the highest score of 0.6840, representing a $23.93\%$ improvement compared to the SDXL Turbo outputs without optimization. Subsequently, GA Empty attained a score of 0.6056, reflecting a $9.37\%$ relative improvement, followed by Promptist with a score of 0.5941 and a $7.64\%$ improvement. Finally, GA Random produced a score of 0.5654 with a $2.45\%$ improvement, whereas Random Search ranked last with a score of 0.5107 and a $-7.47\%$ change.

Based on these findings, it can be concluded that the GA Mutated variant offers more advantages than other GA variations and baselines, including Promptist. Although it did not achieve the highest improvement in terms of aesthetics, the difference was less than 2\% compared to the best result. The most significant achievement was the enhancement in prompt-image alignment: while almost all other methods resulted in a decline in CLIPScore, the GA Mutated variant demonstrated an improvement of 22\%, while Promptist presented a 5.09\% increase. For the GA Mutated, this led to an average fitness improvement of 24\%, which is at least 14\% higher than other GA variants, Promptist, and Random Search. This is evident in the number of prompts "won," in which the optimization method attained the highest fitness score. GA Mutated prevailed in 28 prompts, followed by Promptist with seven and GA Empty with one, while the remaining methods scored zero.

Figures \ref{fig:results_1} and \ref{fig:results_2} illustrate the outputs of the prompt optimization experiment. Each row displays the generated images corresponding to prompts from one of the 12 categories. For the full results, see Figures \ref{fig:results_a1} to \ref{fig:results_a6} in the Appendix.

Through image analysis, it was apparent that GA Mutated and Promptist were the sole methods that consistently preserved semantic similarity with the prompt, as well as with the original non-optimized images. In most instances, the outputs generated by GA Mutated exhibited greater detail than the original outputs, whereas Promptist remained closely aligned with the initial prompt output. Conversely, both GA Random and Random Search frequently devolved into bland scenes, characterized by a recurrent desaturated pastel color palette.

\section{Conclusion and Future Work}
\label{sec:conclusions}

This work demonstrates that evolutionary optimization at the token level constitutes a robust and effective approach for prompt optimization within text-to-image generation, including being an alternative to LLM-based prompt rewriting. By evolving CLIP token vectors with a Genetic Algorithm (GA) and optimizing a joint fitness function that balances aesthetic quality and prompt–image alignment, the proposed approach consistently outperformed Promptist and random search across a diverse subset of Parti Prompts (P2) prompts, achieving up to a 23.93\% improvement in overall fitness. 

Beyond raw performance, the method offers key advantages in terms of model-agnosticism, as it can be applied to other generative models with encoders that perform prompt tokenization. It also benefits from independence from large, biased training datasets because it operates directly on the text encoder’s token space rather than relying on learned linguistic heuristics. These results highlight the potential of evolutionary strategies for exploring conditioning spaces that lie outside conventional human language while preserving semantic intent, suggesting a promising direction for scalable, modular, and controllable prompt optimization in future generative systems. 

Despite these promising results, this work acknowledges several limitations that suggest directions for future research. First, an experimental evaluation was conducted on a relatively small subset of the P2 dataset and was executed solely with SDXL Turbo. Although this model is computationally efficient, it may not fully represent the behavior of larger or multistage diffusion pipelines. Second, the fitness function depends on proxy metrics, namely the LAION Aesthetic Predictor V2 and CLIPScore, which, although widely utilized, are known to exhibit biases and may not accurately reflect human preferences or the requirements of downstream tasks. Third, the current GA configuration and hyperparameters were manually selected and fixed across all prompts, potentially limiting their adaptability to different prompt categories or optimization objectives. 

Therefore, future research should explore larger and more diverse benchmarks, extend the experimental framework to other diffusion architectures, text encoders, and evaluation methods, and investigate adaptive or multi-objective evolutionary strategies that dynamically balance aesthetics, semantic fidelity, and additional constraints such as diversity or robustness. Furthermore, integrating human-in-the-loop evaluation, alternative perceptual metrics, or hybrid approaches that combine evolutionary search with gradient-based or learned optimizers can further enhance the effectiveness and generalization of the token-level prompt optimization.

\paragraph{Disclaimer.} Large language models were used for language editing (grammar, style, and clarity). All technical and scientific content, including claims, experimental design, results, and conclusions, is the responsibility of the authors.

\section*{Acknowledgments}

This work is funded by national funds through FCT – Foundation for Science and Technology, I.P., within the scope of the research unit UID/00326 - Centre for Informatics and Systems of the University of Coimbra, and through the Portuguese Recovery and Resilience Plan (PRR) through project C645008882-00000055, Center for Responsible AI.

\bibliographystyle{unsrtnat}  
\bibliography{references}  

\clearpage
\appendix
\section*{Appendix}
\label{sec:appendix}
\setcounter{figure}{0}\setcounter{table}{0}
\renewcommand{\thefigure}{A\arabic{figure}}
\renewcommand{\thetable}{A\arabic{table}}

\section{Full List of Generated Images}

\begin{center}
\vspace{-2\baselineskip}
  \includegraphics[width=\textwidth]{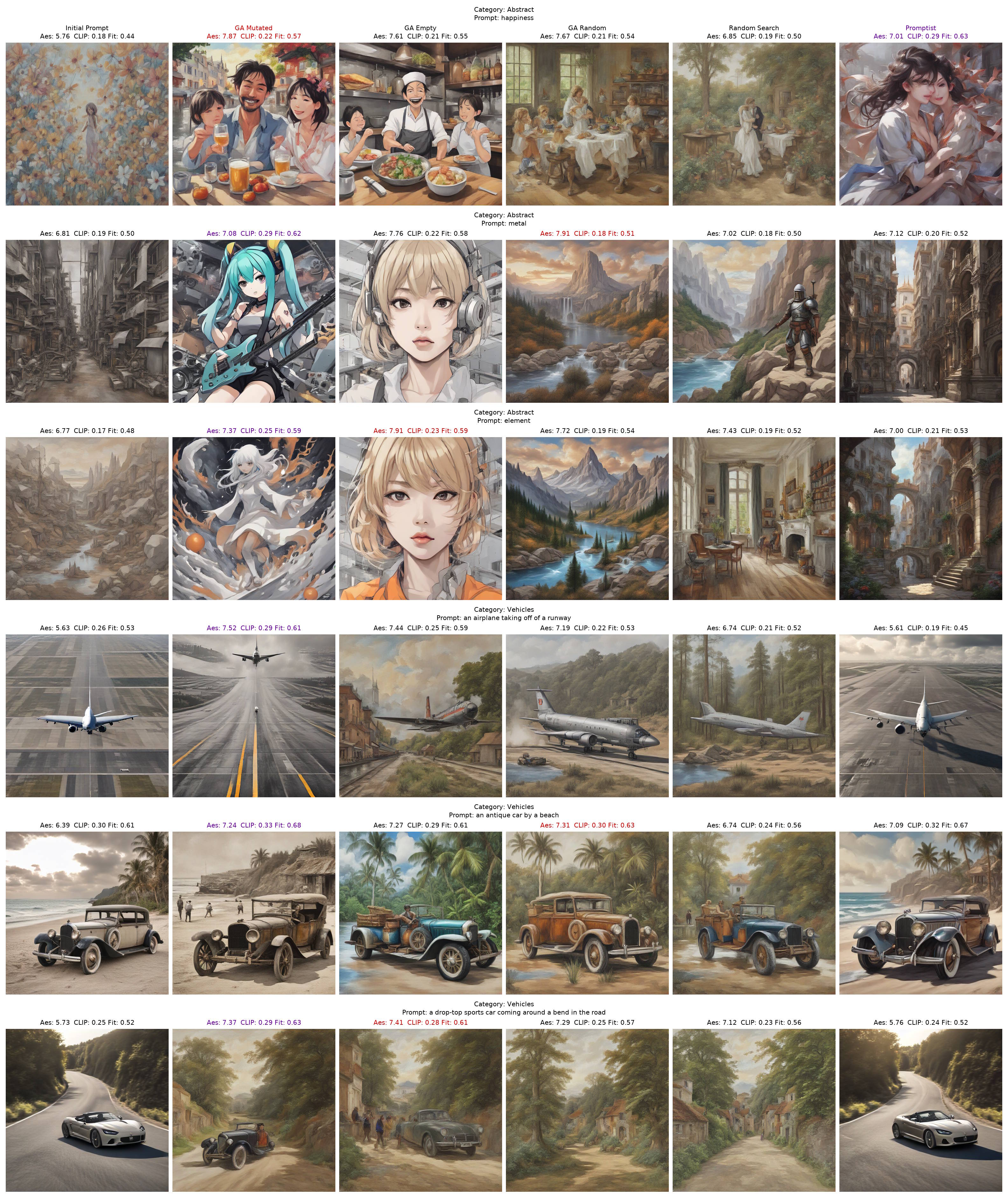}
  \captionof{figure}{Final outputs from baseline SDXL Turbo, GA Mutated, GA Empty, GA Random, Random Search, and Promptist for prompts 1 to 6. Rows correspond to prompts and columns to methods, with aesthetic, CLIP, and fitness scores above each image; purple marks the highest-fitness image, while red or blue mark the best aesthetic or CLIPScore when they do not match the fitness optimum.}
  \label{fig:results_a1}
\end{center}

\begin{center}
  \includegraphics[width=\textwidth]{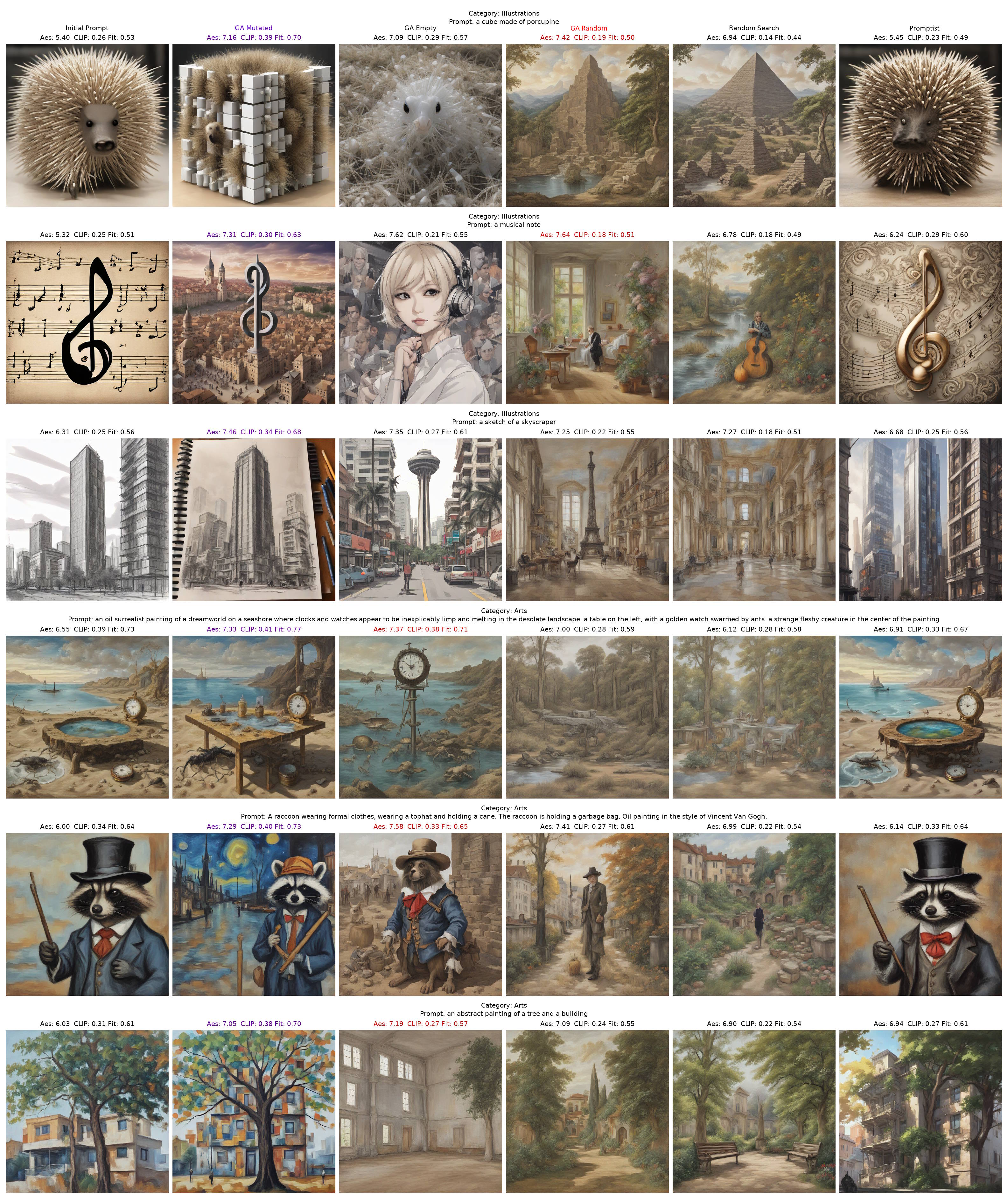}
  \captionof{figure}{Final outputs from baseline SDXL Turbo, GA Mutated, GA Empty, GA Random, Random Search, and Promptist for prompts 7 to 12. Rows correspond to prompts and columns to methods, with aesthetic, CLIP, and fitness scores above each image; purple marks the highest-fitness image, while red or blue mark the best aesthetic or CLIPScore when they do not match the fitness optimum.}
  \label{fig:results_a2}
\end{center}

\begin{center}
  \includegraphics[width=\textwidth]{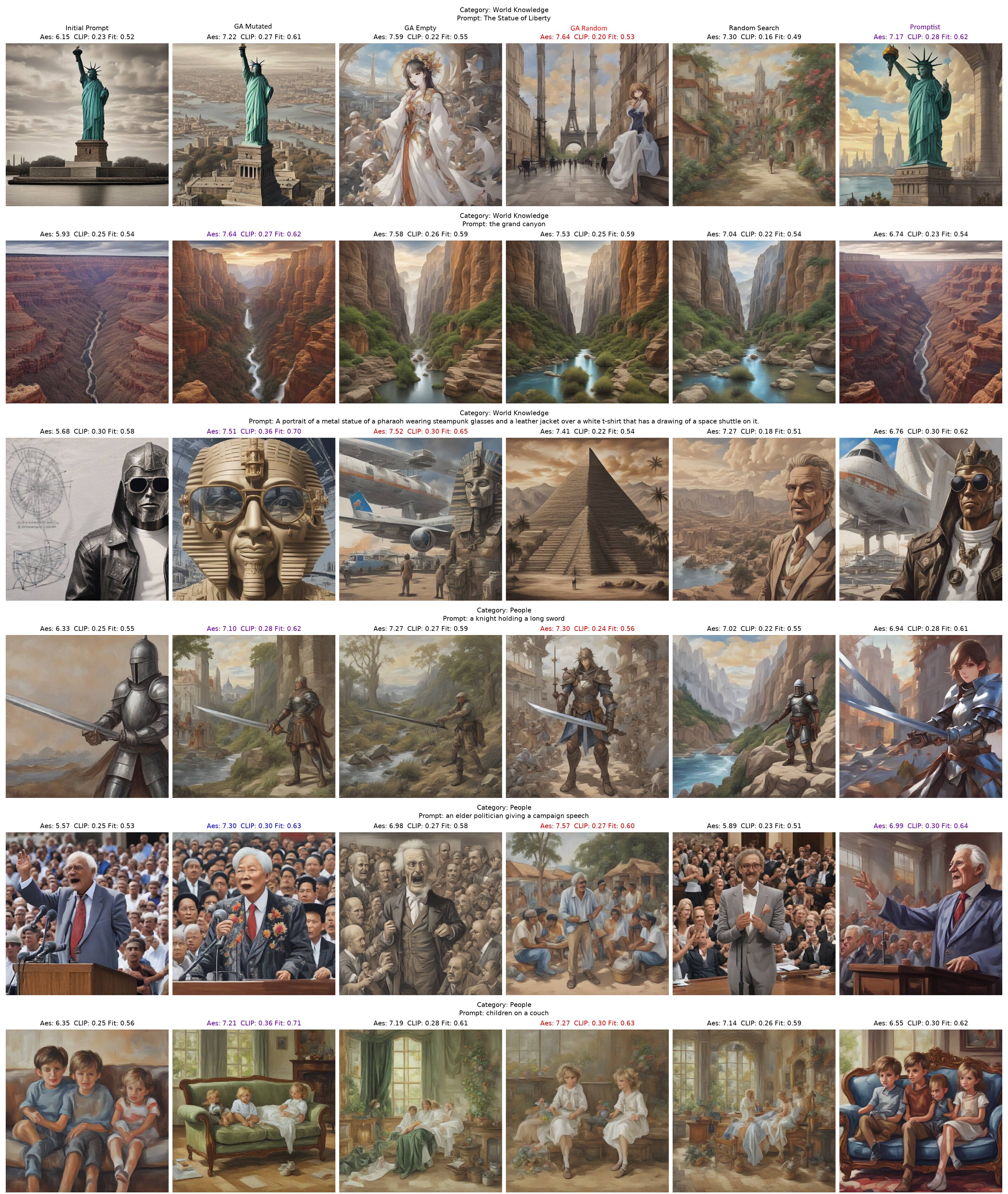}
  \captionof{figure}{Final outputs from baseline SDXL Turbo, GA Mutated, GA Empty, GA Random, Random Search, and Promptist for prompts 13 to 18. Rows correspond to prompts and columns to methods, with aesthetic, CLIP, and fitness scores above each image; purple marks the highest-fitness image, while red or blue mark the best aesthetic or CLIPScore when they do not match the fitness optimum.}
  \label{fig:results_a3}
\end{center}

\begin{center}
  \includegraphics[width=\textwidth]{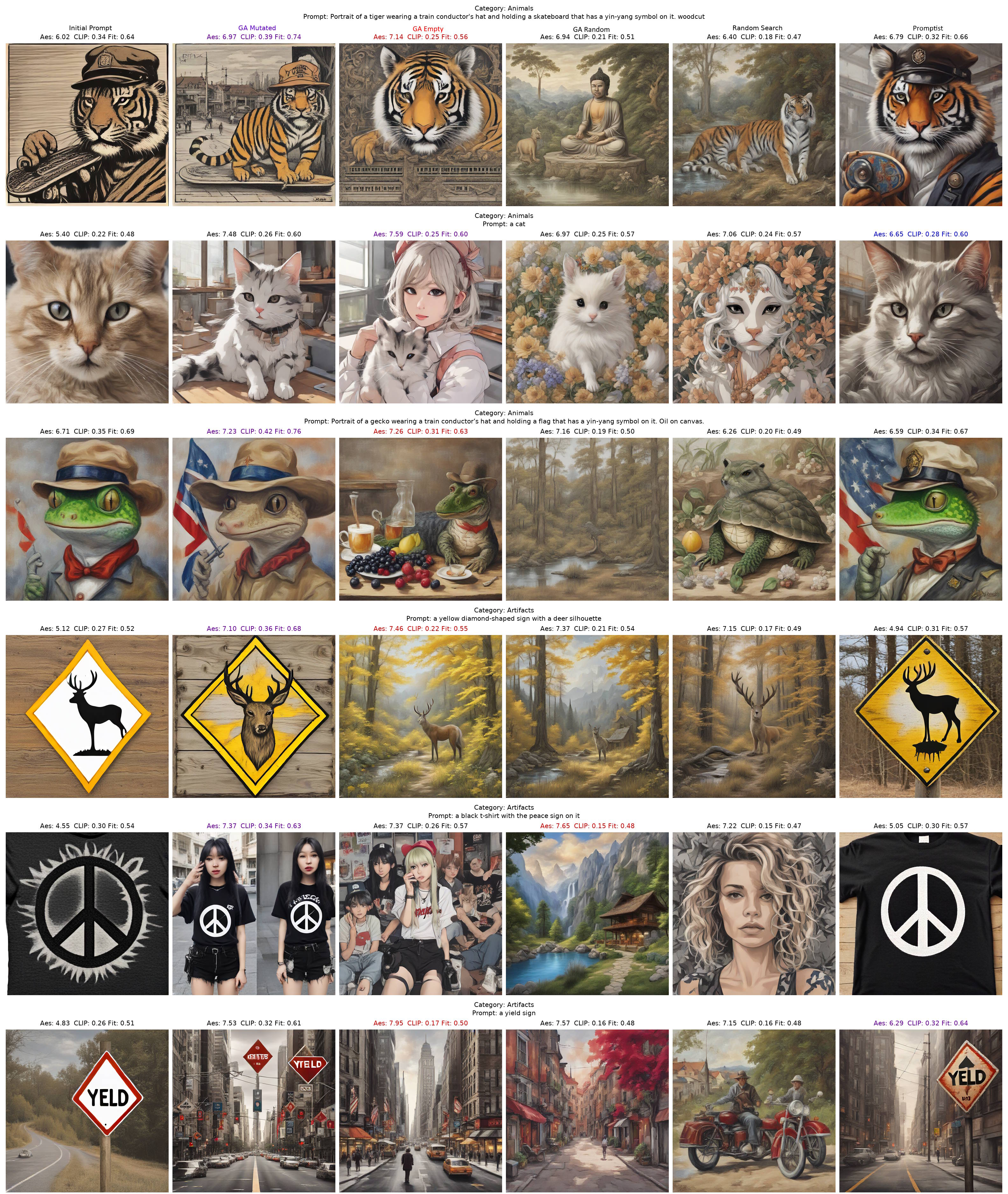}
  \captionof{figure}{Final outputs from baseline SDXL Turbo, GA Mutated, GA Empty, GA Random, Random Search, and Promptist for prompts 19 to 24. Rows correspond to prompts and columns to methods, with aesthetic, CLIP, and fitness scores above each image; purple marks the highest-fitness image, while red or blue mark the best aesthetic or CLIPScore when they do not match the fitness optimum.}
  \label{fig:results_a4}
\end{center}

\begin{center}
  \includegraphics[width=\textwidth]{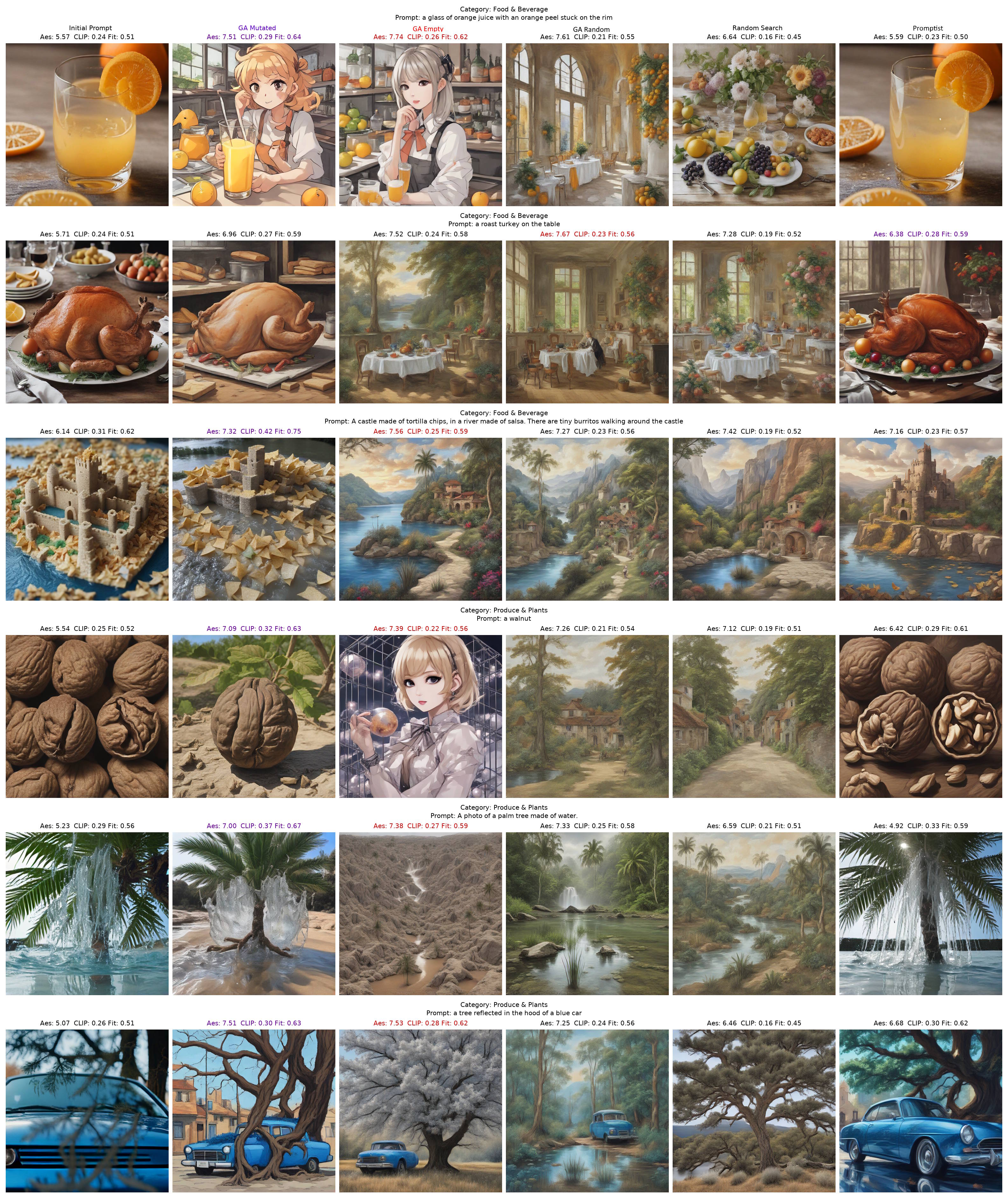}
  \captionof{figure}{Final outputs from baseline SDXL Turbo, GA Mutated, GA Empty, GA Random, Random Search, and Promptist for prompts 25 to 30. Rows correspond to prompts and columns to methods, with aesthetic, CLIP, and fitness scores above each image; purple marks the highest-fitness image, while red or blue mark the best aesthetic or CLIPScore when they do not match the fitness optimum.}
  \label{fig:results_a5}
\end{center}

\begin{center}
  \includegraphics[width=\textwidth]{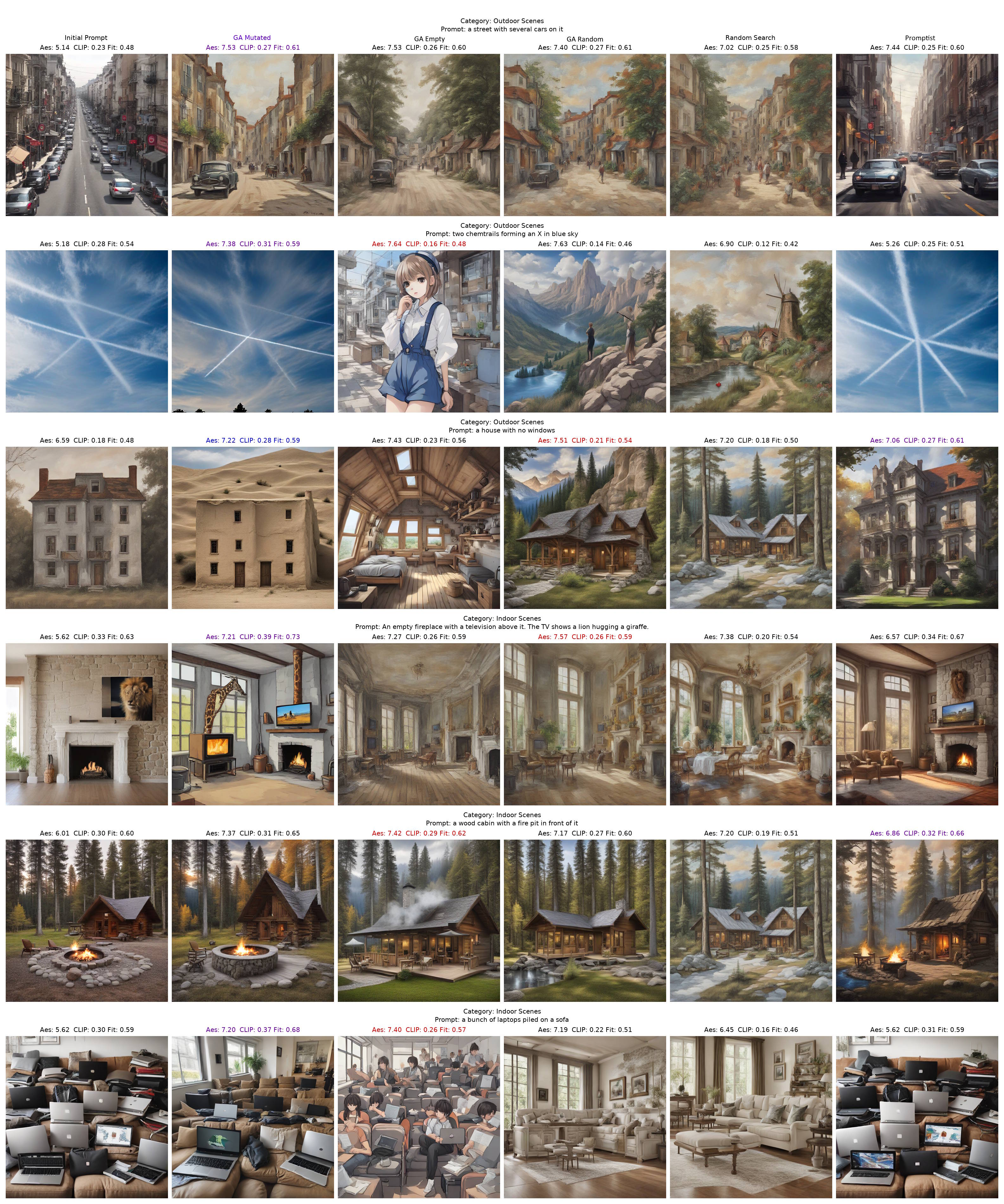}
  \captionof{figure}{Final outputs from baseline SDXL Turbo, GA Mutated, GA Empty, GA Random, Random Search, and Promptist for prompts 31 to 36. Rows correspond to prompts and columns to methods, with aesthetic, CLIP, and fitness scores above each image; purple marks the highest-fitness image, while red or blue mark the best aesthetic or CLIPScore when they do not match the fitness optimum.}
  \label{fig:results_a6}
\end{center}

\end{document}